\documentclass{article}

\PassOptionsToPackage{square,numbers}{natbib}
\usepackage[final]{neurips_2024}
\usepackage{caption}
\usepackage{subcaption}

\usepackage[utf8]{inputenc} % allow utf-8 input
\usepackage[T1]{fontenc}    % use 8-bit T1 fonts
\usepackage{hyperref}       % hyperlinks
\usepackage{url}            % simple URL typesetting
\usepackage{booktabs}       % professional-quality tables
\usepackage{nicefrac}       % compact symbols for 1/2, etc.
\usepackage{microtype}      % microtypography
\usepackage{xcolor}         % colors

\usepackage{graphicx} % Required for inserting images

\usepackage{mathtools} % Improved amsmath
\usepackage{amsfonts}
\usepackage{amsthm} % For the proof environment
\usepackage{thmtools} % Most of what you need for mathematical statements.
\usepackage{thm-restate} % Allows to have proofs separated from the statements
\usepackage{bm} % For bold mathematical fonts

\usepackage[capitalize,noabbrev]{cleveref}

\newtheorem{thm}{Theorem}%[section]
\crefname{thm}{theorem}{theorems}
\Crefname{thm}{Theorem}{Theorems}

\newtheorem{cor}[thm]{Corollary}%[section]
\crefname{cor}{corollary}{corollaries}
\Crefname{cor}{Corollary}{Corollaries}

\crefname{prop}{proposition}{propositions}
\Crefname{prop}{Propostion}{Propositions}

\newtheoremstyle{remark}% name
  {\topsep}% space above
  {\topsep}% space below
  {}% body font
  {}% indent amount
  {\itshape}% theorem head font
  {}% punctuation after theorem head
  {.5em}% space after theorem head
  {\thmname{#1}\thmnumber{ #2}\thmnote{ (#3)}}% theorem head spec
\theoremstyle{remark}
\newtheorem{rmk}[thm]{Remark}%[section]
\crefname{rmk}{remark}{remarks}
\Crefname{rmk}{Remark}{Remarks}

\usepackage{tikz}
\usetikzlibrary{shapes.geometric,positioning,bayesnet,calc}
\usetikzlibrary{arrows,shapes,plotmarks,positioning}
\usetikzlibrary{decorations.markings}
\tikzset{>=stealth'} 
\tikzstyle{graphnode} = 
   [circle,draw=black,minimum size=22pt,text centered,text
     width=22pt,inner sep=0pt] 
\tikzstyle{var}   =[graphnode,fill=white]
\tikzstyle{vardashed}   =[graphnode,draw=gray,fill=white]
\tikzstyle{obs}   =[graphnode,fill=black,text=white]
\tikzstyle{obsgrey}   =[graphnode,draw=white,fill=lightgray,text=black]
\tikzstyle{par}    =[graphnode,draw=white,fill=red,text=black] 
 \tikzstyle{crucial} =[graphnode,draw=white,fill=yellow,text=black] 
\tikzstyle{fac}   =[rectangle,draw=black,fill=black!25,minimum size=5pt]
\tikzstyle{facprior} =[rectangle,draw=black,fill=black,text=white,minimum size=5pt]
\tikzstyle{edge}  =[draw=white,double=black,very thick,-]
\tikzstyle{blueedge}  =[draw=white,double=blue,very thick,-]
\tikzstyle{rededge}  =[draw=white,double=red,very thick,-]
\tikzstyle{prior} =[rectangle, draw=black, fill=black, minimum size=
5pt, inner sep=0pt]
\tikzstyle{dirprior} = [circle, draw=black, fill=black, minimum
size=5pt, inner sep=0pt]
\tikzstyle{dot_node}=[draw=black,fill=black,shape=circle]

\newcommand{\R}{\mathbb{R}}
\newcommand{\E}{\mathbb{E}}

\newcommand{\bX}{\mathbf{X}}
\newcommand{\bW}{\mathbf{W}}
\newcommand{\bx}{\mathbf{x}}
\newcommand{\bmu}{\bm{\mu}}
\newcommand{\bphi}{\bm{\phi}}
\newcommand{\blambda}{\bm{\lambda}}
\newcommand{\bSigma}{\mathbf{\Sigma}}

\newcommand{\YYY}{\mathcal{Y}}

\newcommand{\barbx}{\bar{\bx}}
\newcommand{\barx}{\bar{x}}
\newcommand{\bars}{\bar{s}}

\newcommand{\indep}{\perp \!\!\! \perp} % Conditional independence.
\newcommand{\causal}{\text{causal}}
\newcommand{\anticausal}{\text{anticausal}}

\newcommand{\cdf}{\mathrm{P}}
\newcommand{\pdf}{\mathrm{p}}

\newcommand{\set}[1]{\{#1\}}

\newcommand*\dif{\mathop{}\!\mathrm{d}}

\DeclareMathOperator*{\vecspan}{span}
\newcommand{\Var}{\mathrm{Var}}
\newcommand{\Cov}{\mathrm{Cov}}

\title{Causal vs. Anticausal merging of predictors}

\author{%
  Sergio Hernan Garrido Mejia \\
  Max Planck Institute for Intelligent Systems\\
  Amazon\\
  Tübingen, Germany\\
  \texttt{shgm@tuebingen.mpg.de} \\  
  \And
  Patrick Bl\"obaum \\
  Amazon\\
  Tübingen, Germany\\
  \And
  Bernhard Sch\"olkopf \\
  Max Planck Institute for Intelligent Systems\\
  Tübingen, Germany\\
  \And
  Dominik Janzing \\
  Amazon\\
  Tübingen, Germany\\
}

\begin{document}

\maketitle
\begin{abstract}
We study the differences arising from merging predictors in the causal and anticausal directions using the same data.
In particular we study the asymmetries that arise in a simple model where we merge the predictors using one binary variable as target and two continuous variables as predictors.
We use Causal Maximum Entropy (CMAXENT) as inductive bias to merge the predictors, however, we expect similar differences to hold also when we use other merging methods that take into account asymmetries between cause and effect.
We show that if we observe all bivariate distributions, the CMAXENT solution reduces to a logistic regression in the causal direction and Linear Discriminant Analysis (LDA) in the anticausal direction.
Furthermore, we study how the decision boundaries of these two solutions differ whenever we observe only some of the bivariate distributions implications for Out-Of-Variable (OOV) generalisation.
\end{abstract}

\section{Introduction}

A common problem in machine learning and statistics consists of estimating or combining models or (expert opinions) of a target variable of interest into a single, hopefully better, model \cite{hastie2009elements}.
There are several reasons of why this problem is important.
For example, experts might have access to different data when creating their models, but might not have access to the data available to other experts, while there might be a modeller who can access the expert's opinions and put them together into a single model.
Furthermore, experts might specialise in certain areas of the support of the input space, so that a modeller with access to the expert's opinions could potentially produce a single model exploiting the strengths of each modeller.
This problem is commonly known as ``mixture of experts'', ``expert aggregation'', ``merging of experts'' or ``expert pooling'' \citep{wolpert1992stacked,jordan1994hierarchical,breiman1996bagging,raftery1997bayesian,poole2000inference}.

The merging of experts problem is usually ill-defined, in the sense that there are multiple joint models (that is, models that include \emph{all} covariates) that after marginalisation would render the same prediction as the individual experts (that is, those which include only \emph{some} of the covariates).
This ill-definedness of the problem requires strong inductive biases.
One way to provide this inductive bias in a principled way is through the Maximum Entropy (MAXENT) principle \citep{jaynes1957information}.
In brief, MAXENT suggests finding the distribution with maximum Shannon entropy subject to moment constraints.
This turns out to be the same as choosing the distribution closest to the uniform distribution having the same moments as those given by the constraints.
In \Cref{ssec:MAXENT} we introduce MAXENT and Causal MAXENT (CMAXENT) in more detail, the latter being an extension that allows to include causal information when available \citep{janzing2021causal}.

Most of the research on the merging of predictors focuses on finding a meta-predictor that uses the given models and best fits to the data \cite{yao2022locking}, whereas the focus of the present article is understanding the implications of causal assumptions in the merging of experts instead of aiming for models with best performance.
Furthermore, most of this research focuses on predictors that use the same predicting variables for each of the models.
To the best of our knowledge, the only exception is the random subspace method, notable for being the basis of random forests \citep{ho1998random}.

What if in addition to the different models, a researcher has some causal knowledge of the underlying system?
For example, they could know whether a variable or set of variables used in a particular model are causes or effects of the target variable, potentially changing the resulting predictor of interest.
Causal knowledge produces asymmetries that have been exploited in the past to understand some common machine learning tasks like transfer learning, semi-supervised learning or distribution shift \citep{scholkopf2012causal,weichwald2014causal,janzing2015semi,jin2021causal}.

In the present work, we investigate how including causal knowledge produces asymmetric results when merging predictors.
In particular, we are interested in how solutions to the CMAXENT principle \citep{janzing2021causal} differ when we assume different causal relations for the same data.
That is, we are interested in the asymmetries produced by \emph{causal} assumptions on CMAXENT inferences.
In particular, we will study the differences in the solutions when we assume the causal data generation process (so that the covariates or predictors are causal parents of our target variable) in contrast with the anticausal generation process (so that the predictors are causal children of our target variable).
We are going to study these asymmetries in the case where we do not observe all the variables jointly; one of the differentiating characteristics of this research with respect to other merging of predictors work.

Including the right causal assumptions when merging predictors is relevant, for instance, in the medical domain. 
Suppose we are interested in the presence or absence of a disease, and we have models from hospitals and labs relating risk factors and symptoms to the disease we are interested in.
Combining the predictors would be valuable to predict the disease but it also requires to include the right causal assumptions, if the direction matters for the merging of predictors: risk factors cause diseases and diseases cause symptoms.
The literature of merging of predictors has focused on important aspects of the resulting models like generalisation bounds or speed of estimation but the relation to causality has remained largely unexplored.

Previous approaches have considered the problem of merging of experts using MAXENT \citep{levy1994maximum,myung1996maximum, saerens2004yet}. 
However, such research considers the problem from a purely statistical perspective and does not study the ramifications of different causal assumptions.
In fact, the way they study the aggregation problem is by merging the probability of the outcome given by each expert without regarding how these probabilities were produced.

The main contributions of this article can be summarised as follows
\begin{itemize}
    \item We study the differences in causal and anticausal merging of predictors whenever the inductive bias used to merge the predictors allows causal information to be included.
    \item In particular, we find that CMAXENT with a binary target and continuous covariates, reduces to logistic regression and LDA, two classic classification algorithms, when merging predictors in causal and anticausal directions, respectively.
    \item Furthermore, we study the implications of these asymmetries Out Of Variable (OOV) generalisation whenever we do not observe all the first and second moments as constraints in the CMAXENT problem.
\end{itemize}

The remainder of the paper is organised as follows. 
In \Cref{sec:notationAndPreliminaries} we introduce basic notation and give a brief overview of MAXENT and CMAXENT.
Then, in \Cref{sec:knownPredictorCovariances} we present the optimisation problem in the causal and anticausal direction and give the explicit solutions of the problems, thereby connecting the solutions of CMAXENT to well-known classification algorithms.
In addition, we prove that the decision boundary in the causal and anticausal directions, with full knowledge of the moments (as defined in the section itself) renders equal slopes of the predictors.
In \Cref{sec:partiallyKnownPredictorCovariances} we weaken the assumption of full knowledge of the moments and instead assume knowledge of a subset of the moments in \Cref{sec:knownPredictorCovariances}.
Partial knowledge of the moments have implications for Out Of Variable (OOV) generalisation and resulting in differences in decision boundaries.
We close with \Cref{sec:discussion} with some discussion and concluding remarks.
All the proofs are left to the appendix for the sake of brevity and clarity of the main text.

\section{Notation and preliminaries}
\label{sec:notationAndPreliminaries}

\subsection{Notation}
Let $Y$ be a binary random variable taking values in $\YYY=\set{-1,1}$, and $\bX=\set{X_1,X_2}$ be a pair of continuous variables, so that $x_i\in\R$.
Let $f:\YYY\times \R^2\to \R$ be a measurable function, $\cdf$ a measure on $\YYY\times \R^2$, and $\pdf$ the density of the distribution of a random variable with respect to the Lebesgue measure in the case of real valued random variables, and with respect to the counting measure in the case of discrete random variables.
To be precise, $\pdf(Y,\bX)$ is a density with respect to the product of the Lebesgue measure and the counting measure.
We denote $\E_{\pdf}[f(Y,\bX)]$ the expectation of $f$ with respect to $\pdf$.
We restrict ourselves to the scenario with two continuous variables and one binary outcome given that we can already observe asymmetries in the merging of experts, and can visualise such asymmetries without having to project such space into 2 dimensions.
The results here can be easily generalised into a discrete outcome variable (and indeed we do, in \Cref{th:exponentialFamilyDiscriminantAnalysis}).
Throughout the article we will care about finding a predictor of $Y$ using $\bX$ as covariates.
That is, we are interested in the density $\pdf(Y\mid \bX)$.

\subsection{Maximum Entropy and Causal Maximum Entropy}
\label{ssec:MAXENT}

The Maximum Entropy (MAXENT) principle was born in the statistical mechanics literature as a way to find a distribution consistent with a set of expectation constraints \citep{jaynes1957information}. 
That is, given observed sample averages $\tilde{f}=\frac{1}{N}\sum_{i=1}^Nf(y_i, x_i)$ we find the density $\pdf(Y,\bX)$ so that the expectations with respect to $\pdf(Y,\bX)$ are equal to those observed.

Notice that MAXENT does not attempt to find the `true' distribution of the data, but instead the distribution closest to the uniform distribution so that the expectation constraints are satisfied.
We will see examples of such optimisation problems in subsequent sections.
Using the Lagrange multiplier formalism for constrained optimisation, one can prove that the solution to the MAXENT problem belongs to the exponential family.
The MAXENT distribution and its properties have been studied widely, see \citet{grunwald2004game} and \citet{wainwright2008graphical} and references therein.

In \textbf{Causal MAXENT} (CMAXENT, \citet{janzing2021causal}), the optimisation is performed in an assumed causal order;
that is, we first find the MAXENT distribution of causes and then the Maximum Conditional Entropy of the effects given the inferred distribution of the causes. 
As argued in \cite{sun2006causal} this typically results in distributions that are more plausible for the respective causal direction.
One can think of CMAXENT as usual MAXENT with the distribution of the cause as additional constraint, where the latter has been obtained via separate entropy maximization.

\section{Known predictor covariances}
\label{sec:knownPredictorCovariances}

We will begin by studying the solution of the CMAXENT problem when we observe all the bivariate distributions and summarise them with first and second moments.
The restriction to first and second moments has several reasons: 
First, these simple constraints are already sufficient to explain the interesting asymmetries between causal and anticausal.
Second, including higher order moments makes the problem computationally harder and increases the risk of overfitting on noisy finite sample results.
Last, including more moments decreases the asymmetries between the causal directions.
Mathematically, we have  the following (estimated) expectations and their respective sample averages:
\begin{align}
    &\hat{\E}[Y] =q,\,
    &&\hat{\E}[\bX Y] =\bphi 
    =\begin{bmatrix}
        \phi_1\\
        \phi_2
    \end{bmatrix},\label{eq:knownExpectations1}\\
    &\hat{\E}[\bX] =\barbx 
    =\begin{bmatrix}
        \barx_1\\
        \barx_2
    \end{bmatrix}
    =\begin{bmatrix}
        0\\
        0
    \end{bmatrix},\,
    &&\hat{\E}[\bX\bX^\top] =\bSigma_\bX
    =\begin{pmatrix}
    \bars_1^2   & \bars_{1,2}\\
    \bars_{1,2} & \bars_2^2
    \end{pmatrix},\label{eq:knownExpectations2}
\end{align}
where we assumed the mean of $\bX$ is zero.

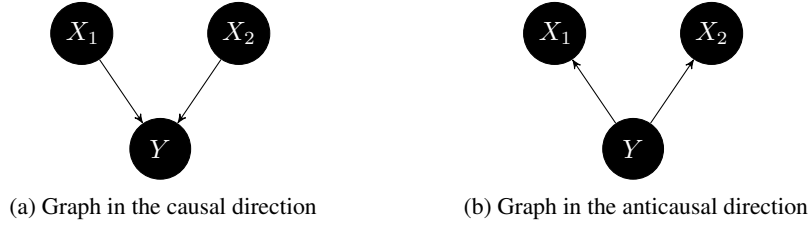
\begin{figure}
     \centering
     \begin{subfigure}[b]{0.45\textwidth}
         \centering
         %!TEX root = CausalityCausalAndAnticausalMergingOfPredictors.tex

\begin{tikzpicture}
\node(Ref) {};
\node[obs,left=.5cm of Ref] (X1) {$X_1$};
\node[obs,right=.5cm of Ref] (X2) {$X_2$};
\node[obs,below=1cm of Ref] (Y) {$Y$};

\edge{X1}{Y};
\edge{X2}{Y};

\end{tikzpicture}
         \caption{Graph in the causal direction}
         \label{fig:graphCausalDirection}
     \end{subfigure}%\hfill
     \begin{subfigure}[b]{0.45\textwidth}
         \centering
         %!TEX root = CausalityCausalAndAnticausalMergingOfPredictors.tex

\begin{tikzpicture}
\node(Ref) {};
\node[obs,left=.5cm of Ref] (X1) {$X_1$};
\node[obs,right=.5cm of Ref] (X2) {$X_2$};
\node[obs,below=1cm of Ref] (Y) {$Y$};

\edge{Y}{X1};
\edge{Y}{X2};

\end{tikzpicture}
         \caption{Graph in the anticausal direction}
         \label{fig:graphAnticausalDirection}
     \end{subfigure}
    \caption{Causal graphs analysed throughout the article}
    \label{fig:causalGraphs}
\end{figure}

\subsection{The causal direction}
\label{ssec:causalDirection}

Consider the causal graph in \Cref{fig:graphCausalDirection} and the expectations given in \Cref{eq:knownExpectations1,eq:knownExpectations2}.
As mentioned on \Cref{ssec:MAXENT}, CMAXENT suggests finding the density $\pdf(\bX)$ with maximum entropy consistent with the first and second moments of $\bX$, and then finding the density $\pdf(Y\mid \bX)$ with maximum entropy, with the estimated $\pdf(\bX)$ and the moments that involve $Y$ as constraints.

These steps can be summarised in the following optimisation problems.
First, for $\pdf(\bX)$,
\begin{align}\label{eq:causalMarginalOptimisastion}
\begin{split}
    \max_{\pdf(\bx)}& \quad H(\bX) = - \int_{\R^2}\pdf(\bx)\log \pdf(\bx)\dif\bx\\
    \text{s.t.}\quad  & \quad  \E[X_i]= \bar{x}_{i},\, \text{with $i\in\{1,2\}$}\\
    & \quad  \E[X_i^2] = \bar{s}^{2}_{i},\, \text{with $i\in\{1,2\}$}\\
    & \quad  \E[X_1X_2] = \bar{s}_{1,2}\\
    &\quad \int_{\R^2}\pdf(\bx)\dif\bx = 1.
\end{split}
\end{align}
On the other hand, the Maximum Conditional Entropy  optimisation problem is as follows
\begin{align}\label{eq:causalConditionalOptimisastion}
\begin{split}
    \max_{\pdf(y\mid \bx)}& \quad H(Y\mid \bX) = - \int_{\R^2}\sum_{y} \pdf(y\mid\bx)\pdf(\bx)\log \pdf(y\mid\bx)\dif\bx\\
    \text{s.t.}& \quad  \E[YX_i] = \phi_{i},\, \text{with $i\in\{1,2\}$}\\
    &\quad \E[Y] = q\\
    &\quad \sum_{y} \pdf(y\mid\bx) = 1,\quad \text{for each $\bx$}.
\end{split}    
\end{align}
Where $\pdf(\bX)$ is the one we found by solving \Cref{eq:causalMarginalOptimisastion}.

\begin{restatable}[Resulting predictor in the causal direction]{prop}{causalDirection}\label{th:causalDirection}
Using the Lagrange multiplier formalism for the optimisation problems in \Cref{eq:causalMarginalOptimisastion,eq:causalConditionalOptimisastion} we obtain: (i) a multivariate Gaussian distribution for $P(\bX)$, and (ii) the density of $Y$ conditioned on $\bX$ given by
\begin{align}
    \pdf_\lambda(y \mid x_{1}, x_{2}) &= \exp\left(
    \lambda_0y + 
    \lambda_{1}yx_1 + 
    \lambda_{2}yx_2 + \alpha(x_{1}, x_{2})\right)\\
    \alpha(x_{1}, x_{2}) &= \log\sum_{y}\exp\left(
    \lambda_0y + 
    \lambda_{1}yx_1 + 
    \lambda_{2}yx_2\right),
\end{align}
where $\alpha(\bx)$ is a normalising constant.

The density can be written as
\begin{align}\label{eq:causalConditionalMaxent}
    \pdf_\lambda(y=1\mid x_{1},x_{2}) &= \frac{1}{2}
    (1+\tanh(\lambda_0 + \lambda_{1}x_1 + \lambda_{2}x_2)).
\end{align}
\end{restatable}
The proof of this result can be found in \citep[Section 3.1 and 3.2]{janzing2009distinguishing}.
\begin{rmk}
    Notice that \Cref{eq:causalConditionalMaxent}, our predictor of interest, is just a rescaled version of a sigmoid function.
    That is, we can estimate $\pdf(Y\mid\bX)$ with a logistic regression.
    A similar observation was done in \cite{friedman2000additive} in the context of using an exponential loss for boosting.
    This relation was further explored in \cite{lebanon2001boosting}, where a more direct relation to maximum likelihood and exponential families was established.
    In \Cref{sec:discussion} we discuss how these results in the statistical literature can be interpreted as making causal assumptions about the relation between the predictor and target variables.
\end{rmk}

\subsection{The anticausal direction}
\label{ssec:anticausalDirection}

Now consider the graph in \Cref{fig:graphAnticausalDirection}.
In this scenario, covariates of our predictor of interest are the effects of our target variable.
Following the CMAXENT principle, we first find the density $\pdf(Y)$ with maximum entropy and is consistent with first moment of $Y$ and then find the density $\pdf(\bX\mid Y)$ with maximum conditional entropy consistent with the moments that involve $\bX$ and $\pdf(Y)$ found in the previous step.
After this two-step process, we are left with the joint density $\pdf(Y,\bX)$ from which we can derive a predictor of $Y$, $\pdf(Y\mid\bX)$ using Bayes' Theorem (\Cref{ssec:predictorAnticausal}).
The whole procedure can be summarised with the following optimisation problems.
For the cause, we have
\begin{align}
\begin{split}
    \max_{\pdf(y)}& \quad H(Y) = - \sum_{y} \pdf(y)\log \pdf(y)\label{eq:anticausalMarginalOptimisation}\\
    \text{s.t.}& \quad  \E[Y] = q\\
    &\quad \sum_{y} \pdf(y) = 1.
\end{split}    
\end{align}
And for the effects,
\begin{align}\label{eq:anticausalConditionalOptimisation}
\begin{split}
    \max_{\pdf(\bx\mid y)}& \quad H(\bX\mid Y) = - \int_{\R^2}\sum_{y} \pdf(\bx\mid\ y)\pdf(y)\log \pdf(\bx\mid y)d\bx\\
    \text{s.t.}& \quad  \E[YX_i] = \phi_{i},\, \text{with $i\in\{1,2\}$}\\
    & \quad \E[X_i] = \bar{x}_{i},\, \text{with $i\in\{1,2\}$}\\
    & \quad  \E[X_i^2] = \bar{s}^{2}_{i},\, \text{with $i\in\{1,2\}$}\\
    & \quad  \E[X_1X_2] = \bar{s}_{1,2}\\
    & \quad \int_{\R^2} \pdf(\bx\mid y)d\bx = 1,\quad \text{for each $y$}.
\end{split}
\end{align}
\begin{restatable}[Resulting predictor in the anticausal direction]{prop}{anticausalDirection}\label{th:anticausalDirection}
Using the Lagrange multiplier formalism for the optimisation problems in \Cref{eq:anticausalMarginalOptimisation,eq:anticausalConditionalOptimisation}, 
we obtain a Bernoulli distribution for $Y$ with $\pdf(y=1)=q$, and $\pdf_\lambda(\bx\mid y)$ given by
\begin{align}\label{eq:exponentialAnticausal}
    \begin{split}
        \pdf_\lambda(\bx \mid y) &= \exp[\lambda_{1}yx_{1} + \lambda_{2}yx_{2} + \lambda_{3}x_{1} + \lambda_{4}x_{2}\\ 
        &\quad\quad\quad + \lambda_{5}x^{2}_{1} + \lambda_{6}x^{2}_{2} + \lambda_{7}x_{1}x_{2} + \beta(y)]
    \end{split}\\
    &= \exp\left[\sum_{k}\lambda_{k}h_{k}(\bx, y) + \beta(y)\right]\\
    \beta(y) &= \log\int_{\R^2} \exp\left[\sum_{k}\lambda_{k}h_{k}(\bx, y)\right] d\bx,
\end{align}
where $h_k$ are the different functions for which we have the sample averages.
The density $\pdf_\lambda(\bX\mid Y)$ is a mixture of multivariate Gaussian distributions.
Both components $\pdf_\lambda(\bX\mid y=-1)$ and $\pdf_\lambda(\bX\mid y=1)$ have the same covariance matrix.
\end{restatable}
For the following sections, we introduce the following notation for the expectations of the mixture of Gaussians. 
\begin{align}\label{eq:conditionalExpectations}
    &\E[\bX\mid y] =\bmu_y = 
    \begin{bmatrix}
    \mu_{y, 1}\\
    \mu_{y, 2}
    \end{bmatrix},
    &&\E[\bX\bX^\top\mid Y] =\bSigma_{\bX\mid Y}
\end{align}
In addition, we will include the subscripts ``causal'' and ``anticausal'' where it might be ambiguous (e.g., $\bSigma_{\bX\mid Y, \causal}$ represents the conditional covariance in the causal scenario and $\bSigma_{\bX\mid Y,\anticausal}$ in the anticausal scenario).
As mentioned in \Cref{th:anticausalDirection}, the conditional covariance $\Sigma_{\bX\mid Y}$ is the same for both values of $y$.
However, we keep the conditional notation to distinguish it from the \emph{marginal} covariance of $\bX$, $\bSigma_\bX$ introduced in \Cref{eq:knownExpectations2}. 
In \Cref{app:relationGaussianConstraints} we derive the conditional expectations in $\Cref{eq:conditionalExpectations}$ and the marginal expectations used as constraints.
\begin{rmk}\label{rmk:marginalAndConditionalCovariance}
Even though the causal graph in the anticausal direction implies that the conditional covariance $\Sigma_{\bX Y}$ is diagonal, the CMAXENT solution does not result in a diagonal conditional covariance.
This is true because of the constraints in  \Cref{eq:knownExpectations1,eq:knownExpectations2} and the law of total covariance.
Note that the CMAXENT distribution is not necessarily Markov relative to the given DAG. 
As shown in \cite[Section 5]{janzing2021causal}, CMAXENT only provides the best guess and may therefore mix over different Markovian distributions such that the result is no longer Markovian.
\end{rmk}
This relation between the marginal and conditional expectations will be essential in the subsequent sections where we explore the difference in the decision boundaries of the two resulting predictors of $Y$.

\subsection{The predictor of $Y$ in the anticausal direction}
\label{ssec:predictorAnticausal}

Recall that our main goal is to produce a predictor of $Y$ as a function of the covariates $\bX$.
In \Cref{ssec:causalDirection} we obtain the predictor of $Y$ directly as a result of the CMAXENT principle, given that the predictor is already in the direction of the causal mechanism.
On the other hand, in \Cref{ssec:anticausalDirection}, we have to derive the predictor of $Y$ using the found conditional distributions and Bayes' rule.
The main result of this section is that with the constraints we have used, CMAXENT in the anticausal direction is equivalent to Linear Discriminant Analysis \citep[Section 4.3]{hastie2009elements}.
Furthermore, we generalise this result to Quadratic Discriminant Analysis, and to an exponential family version of discriminant analysis.

\begin{restatable}[Predictor of $Y$ using Bayes' rule]{thm}{bayesPredictor}\label{th:bayesPredictor}
Using the results from \Cref{th:anticausalDirection}, the density $\pdf_\lambda(Y=y\mid\bX)$ is the ratio of the product of the Gaussian component with $\pdf_\lambda(Y=y)$ and the mixture of Gaussians resulting from \Cref{th:anticausalDirection}.
Minimising the expected 0-1 loss, the optimal decision rule arising from this density is equivalent to Linear Discriminant Analysis (LDA).
\end{restatable}

\begin{cor}[Quadratic Discriminant Analysis (QDA)]
Quadratic Discriminant Analysis can be interpreted as CMAXENT in the anticausal direction. 
This is achieved by replacing 
\begin{align}
    \E[X_i^2] = \bar{s}_{i}^2\, \text{ with $i\in\{1,2\}$, and }\, \E[X_1X_2] = \bar{s}_{1,2}.
\end{align}
in \Cref{eq:anticausalConditionalOptimisation} with the following constraints:
\begin{align}
    \E[X_i^2\mid y] = \bar{s}_{i,y}^2\, \text{ with $i\in\{1,2\}$, and }\, \E[X_1X_2\mid y] = \bar{s}_{1,2,y}.
\end{align}
\end{cor}

We will now extend this idea, where instead of modelling $\pdf(\bX)$ as a mixture of Multivariate Gaussians (with equal covariance in LDA or unequal covariance in QDA), $\pdf(\bX)$ now becomes a mixture of distributions, each coming from an exponential family of distributions corresponding to a more general set of constraints.
\begin{cor}[Exponential family discriminant analysis]\label{th:exponentialFamilyDiscriminantAnalysis}
Let $f_i$ be an arbitrary measurable function and $\tilde{f}$ its corresponding sample average.
In the general case where $Y$ is a discrete variable and we have $d$ covariates $\bX$ in the anticausal direction, the CMAXENT problem with constraints of the form:
\begin{align}
    \E[f_i(\bX)\mid y] = \tilde{f}_{i,y},
\end{align}
where $\tilde{f}_{i,y}$ are the sample averages of $f_i$ for a specific $y$ as in \Cref{ssec:MAXENT}, results in $\pdf_\lambda(\bX\mid Y)$ being a mixture of exponential family distributions which then can be inverted (using Bayes' rule) to a predictor of $Y$.
\end{cor}

\begin{rmk}
In the previous corollary, the functions $f_i$ can be constant on any of the variables in $\bX$.
\end{rmk}

This idea has been extended to use kernels as a way to map $\bX$ into more complex feature spaces.
The resulting algorithm is called Kernel Fisher discriminant analysis \cite{mika1999fisher,roth1999nonlinear,ghojogh2019fisher}.

\subsection{The geometry of the decision boundaries}
\label{ssec:geometryDecisionBoundaries}

\citet[Chapter 4.4.5]{hastie2009elements} conclude that the log-posterior odds of the logistic regression and LDA are both linear in $\bx$, but with different parameters defining the linear relation.
In this section, we revisit these results in more detail and explore whether the CMAXENT solution in causal direction differs from the solution in anticausal direction. 
From a statistical decision theory perspective, the log-posterior odds correspond to the Maximum A Posteriori (MAP) rule, the optimal decision boundary of a classifier when minimising the expected 0-1 loss \cite[Ch. 4.3.3]{berger2013statistical}.

\begin{restatable}[Normal vector to the decision boundaries in causal and anticausal direction]{prop}{decisionBoundaryCMAXENT}\label{th:decisionBoundaryCMAXENT}
Under the 0-1 loss, the normal vector to the decision boundary of the CMAXENT predictor is proportional to
\begin{enumerate}\centering
    \item $\bSigma_{\bX,\causal}^{-1}\bphi$ in the causal direction.\label{th:decisionBoundaryCausal}
    \item $\bSigma_{\bX\mid Y, \anticausal}^{-1}\bphi$ in the anticausal direction.\label{th:decisionBoundaryAnticausal}
\end{enumerate}
\end{restatable}

We will now prove that in the case where we know all the expectations in \Cref{eq:knownExpectations1,eq:knownExpectations2}, the slope of the decision boundaries in causal and anticausal direction are the same.

\begin{restatable}[Slope of the decision boundary is the same in causal and anticausal direction]{thm}{equalityOfSlopes}\label{th:equalityOfSlopes}
Using the constraints in \Cref{eq:knownExpectations1,eq:knownExpectations2}, the slope of $\pdf_\lambda(Y\mid \bX)$ inferred using CMAXENT is the same in causal and anticausal direction.
\end{restatable}

Although it might seem from the above result that there is no asymmetry between the logistic regression and LDA even when we include causal information, this is not entirely true.
To begin with, the decision boundaries may be unequal although they are parallel, but more importantly learning the parameters of certain model might be easier.
In the next section we explore the differences that persist even under the light of \Cref{th:equalityOfSlopes}.

\subsection{What are the differences?}

In the previous sections we found that the slopes of the decision boundary of CMAXENT in both the causal and anticausal direction are linear and agree, whenever we have the first and second moments as in \Cref{eq:knownExpectations1,eq:knownExpectations2}.
This implies that, if the test data will come from the same distribution as the training data, either algorithm will work equally well. 
Previous research has studied the advantages and disadvantages \cite[Chapter 4.4.5]{hastie2009elements,rubinstein1997discriminative} of each method and their properties such as asymptotic relative efficiency \cite{efron1975efficiency}, parameter bias \cite{halperin1971estimation}, asymptotic error under label noise \cite{bi2010efficiency} and online learning performance \cite{banerjee2007analysis}.
All of these analyses base their results on the fact that the logistic regression does not make an assumption on how the covariates $\bX$ are distributed, whereas LDA does.

An alternative way of viewing this distinction is through the lens of generative and discriminative models.
LDA is a generative model since it models both covariates and target variable and logistic regression only models the target as a function of the input.
\citet{ng2001discriminative} analyse the difference in efficiency between the Naive Bayes algorithm (a generative model similar to LDA) and logistic regression, and find that both models have regimes in which they perform better than the other.
Using the same models, \citet{blobaum2015discriminative} and data from \cite{scholkopf2012causal}, find empirically that generative models perform better in anticausal than in causal direction.

\section{Partially known covariances}
\label{sec:partiallyKnownPredictorCovariances}

In this section we explore variations of the solution of the CMAXENT solution in causal and anticausal direction when some of the sample averages are not known. 
In \Cref{ssec:unknownPredictorTarget} we explore the case where the covariance between a particular predictor and the target is not known, and in \Cref{ssec:unknownPredictorPredictor} the case where we do not know the covariance between the predictors.
In both cases we will see that the models we can infer (that is, $\pdf(Y\mid \bX)$) with CMAXENT will depend on the underlying causal assumptions.

\subsection{Unknown predictor-target covariance}
\label{ssec:unknownPredictorTarget}

Without loss of generality, suppose we do not have the sample covariance between $X_2$ and $Y$, that is, we do not know $\phi_2$ in \cref{eq:knownExpectations1}.

In the \textbf{causal direction}, the CMAXENT solution of the distribution of the causes $\bX$ will still be a multivariate normal distribution with expectations given by the constraints relating $\bX$.
The conditional density of the effects is the logistic-like regression of \cref{eq:causalConditionalMaxent}, however, $\lambda_2$ will be 0, as this is the parameter corresponding to the covariance between $X_2$ and $Y$.
In other words, $X_2$ becomes irrelevant in the estimation of our target predictor.

In the \textbf{anticausal direction}, the distribution of the cause $Y$ is unchanged because $\cdf(Y)$ is determined by the constraints and thus does not depend on $\phi_2$.
However, using the fact that the Gaussian distribution maximises the entropy over all distributions with the same variance \citep[Theorem 8.6.5.]{cover2006elements},
we can derive a bound on $\phi_2$.
We use the entropy of the Gaussian distribution because we do not know a closed form expression for the conditional covariance of $\pdf(Y\mid\bX)$ as given by \Cref{th:bayesPredictor}.
 \begin{restatable}[Bounds on unknown covariance between predictor and target]{prop}{covarianceBound}\label{th:covarianceBound}
Assuming the causal graph in \Cref{fig:graphAnticausalDirection} and we do not know $\phi_2$, an upper bound for $\phi_2$ is given by:
\begin{align}
    \frac{q(1-q)\bars_{1,2}\phi_1}{q(1-q)\bars_1^2-\phi_1^2}.
\end{align}
\end{restatable}
The bound in \Cref{th:covarianceBound} is found by differentiating the determinant of the conditional covariance (to which the differential entropy of the multivariate Gaussian is proportional to) with respect to the unknown covariance, $\phi_2$ in this case, and finding the value for $\phi_2$ for which this derivative is 0.

 The implication of the previous result is that even when we have not observed any joint data between $X_2$ and $Y$ we can still build a model of $Y$ that depends on $X_2$, as long as we can assume the data generation process is anticausal.
 We can consider this an instance of Out Of Variable (OOV) generalisation studied in \cite{janzing2018merging,guo2024out}, 
 where we can exploit causal information and partial data to make models including variables that were never observed jointly with the target.

\subsection{Unknown predictor covariance}
\label{ssec:unknownPredictorPredictor}

Now suppose we observe all the sample averages in \Cref{eq:knownExpectations1,eq:knownExpectations2} but we do not observe $\bars_{1,2}$.

In the \textbf{causal} direction this implies that the multivariate Gaussian distribution resulting from the MAXENT problem on $\bX$ is diagonal, that is, $\bX$ are marginally independent.
The exponential form of $\pdf_\lambda(Y\mid X)$ does not change, as we still observed $q$ and $\bphi$, nevertheless, the parameters of the exponential family do change, as the density of $\bX$ changed so that the resulting $\pdf_\lambda(Y\mid \bX)$ needs to adapt in order to match $\bphi$.

Now we will explore the \textbf{anticausal} case.
We will proceed as in \Cref{ssec:causalDirection,ssec:anticausalDirection}.
First we find $\pdf(Y)$ by maximising the entropy subject to the empirical average of $Y$, which is trivial because $\pdf(Y)$ is already determined by its moments, and then we find $\pdf(\bX\mid Y)$ subject to all the moments in \Cref{eq:knownExpectations1,eq:knownExpectations2} with the exception of $\bars_{1,2}$.
We obtain the following result from solving the CMAXENT optimisation problem
\begin{restatable}[Diagonal conditional covariance in the anticausal direction with unknown predictor covariance]{prop}{unknownCovarianceAnticausalDiagonalConditionalCovariance}\label{th:unknownCovarianceAnticausalDiagonalConditionalCovariance}
The density $\pdf(\bX\mid Y)$ that maximises the conditional entropy subject to the following constraints:
\begin{align}
    \hat{\E}[\bX Y]
    =\begin{bmatrix}
        \phi_1\\
        \phi_2
    \end{bmatrix},\quad
    \hat{\E}[\bX]
    =\begin{bmatrix}
        0\\
        0
    \end{bmatrix},\quad
    \hat{\E}[X_1^2] 
    =\bars_1^2,\quad
    \hat{\E}[X_2^2] 
    =\bars_2^2,
\end{align}
and $\pdf(Y)$ inferred on the first step of CMAXENT, is independent after choosing a value of $y$;
that is, $\bX$ is conditionally independent given $Y$.
\end{restatable}
\begin{rmk}
    This result is reassuring given that under these moment constraints, $\pdf(\bX\mid y)$ turns out to be Markov relative to the DAG in the anticausal direction.
    Contrary to \Cref{rmk:marginalAndConditionalCovariance}, where we concluded that CMAXENT is not always Markov relative to a DAG.
\end{rmk}

In \Cref{app:unknownPredictorCovariance} we derive the slopes of the decision boundaries in the causal and anticausal direction when we do not know the covariance between the predictors.
We also find necessary and sufficient conditions for which the slopes are the same.
From this simple example, we have learned the following:
in causal direction, our inductive bias tells us that the covariates are not correlated and hence, the decision boundary depends only on the marginal variance of each $\bX_i$ and the covariance between $Y$ and $\bX$. 
In the anticausal direction, CMAXENT infers $\bX_1$ and $\bX_2$ to be marginally correlated because they need to be conditionally independent (this fact can proved using the law of total covariance).
Hence, the marginal covariance of $\bX$, $\bSigma_\bX$ is different in both scenarios.
This is something we did not observe in the case with full information (\Cref{sec:knownPredictorCovariances}).

\begin{figure}[ht!]
  \centering
  \includegraphics[]{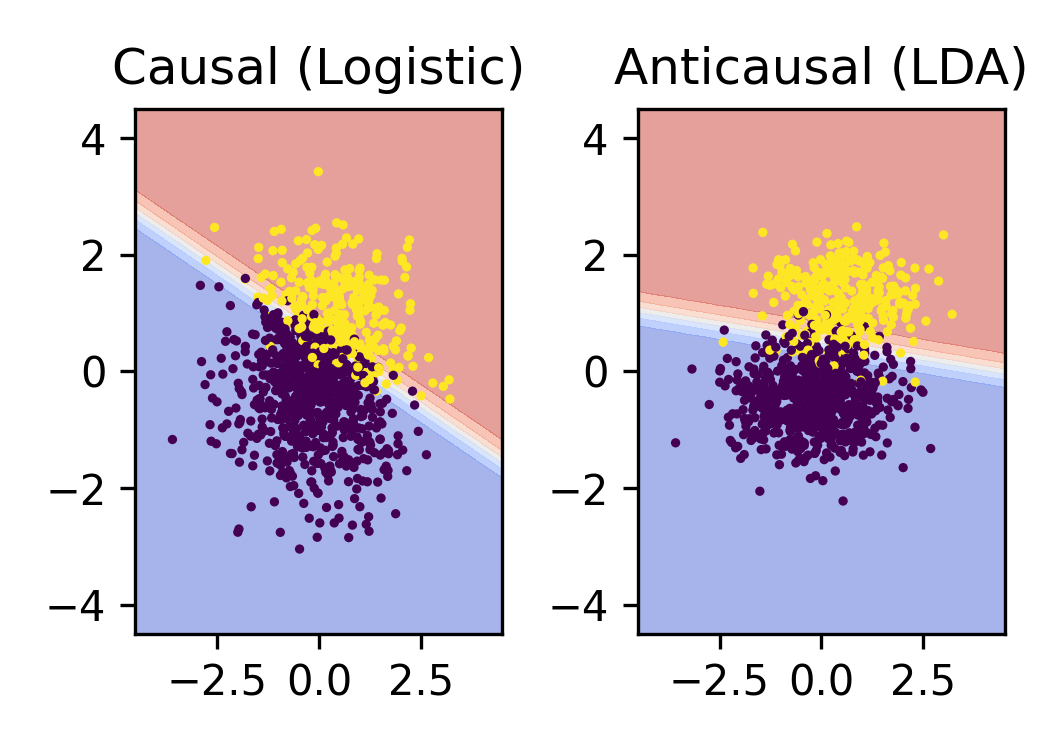}
  \caption{Decision boundaries of the solution of CMAXENT in the causal (left) and anticausal (right) direction when we do not have the covariance between the predictor variables  $\bars_{1,2}$.}
  \label{fig:unknownPredictorCovariance}
\end{figure}

In addition, we derive the expressions of the decision boundaries in the causal and anticausal direction (see \Cref{app:unknownPredictorCovariance}).
That is, as proved in \Cref{th:decisionBoundaryCMAXENT}, we have that the decision boundaries of the predictors in causal and anticausal direction will differ \emph{with the same moments, but different causal assumptions}.
In \Cref{fig:unknownPredictorCovariance}, we showcase this phenomenon with synthetic data.

\section{Discussion}
\label{sec:discussion}

In this article we have studied the differences arising from merging of predictors in the causal and anticausal directions.
In particular, we have studied a simple case with a binary target variable and two continuous variables.
Although in this simple example we have already found connections with classical classification algorithms, and differences in the solutions in causal and anticausal direction, the example can be easily extended to more covariates (where the resulting distribution of the covariates would be a $d$-dimensional Gaussian instead of bivarate Gaussian), a discrete target variable (as in \Cref{th:exponentialFamilyDiscriminantAnalysis}), and a causal graph that contains both parents and children as predictors.

As stated at the end of \Cref{ssec:causalDirection}, the relation between merging of experts and logistic regression has been explored in the past.
\citet{friedman2000additive} interpret the solution to the AdaBoost procedureas as additive logistic regression. 
They arrive at this interpretation starting from an exponential loss function.
They then propose likelihood based estimator of the AdaBoost procedure.
Thus, since our results align with those in  \citet{friedman2000additive}, we give yet another interpretation of AdaBoost as the solution of the merging of experts in causal direction using the CMAXENT principle.

Even though we have used CMAXENT as inductive bias to merge the predictors throughout the article, we believe that the asymmetries we found here (in particular, the geometry of the decision boundaries) would hold when using any other inductive bias that allows causal information to be included.
Whatever method we use to merge predictors,  the following asymmetry seems natural:
In {\it anticausal} direction we try to {\it explain} correlations between $X_1,X_2$ as a result of $Y$ influencing both components.
In {\it causal} direction, correlations between $X_1$ and $X_2$ do not tell us anything about the relation between $X$ and $Y$, following the principle of Independent Mechanisms (see \cite{peters2017elements} for an overview and \cite{guo2023causal} for a recent Bayesian view).

The previous observation is useful in straightforward scenarios where we are merging data from different sources for a supervised learning task, say datasets with overlapping variables, or datasets produced from different experimental conditions (also called environments);
but also in cases where the merging of data is more subtle, for example, in federated learning where the notion of horizontal and vertical federated learning \citep{zhang2021survey} coincides precisely with the data sources described above but where causality is underexplored.

\begin{ack}
We thank William R. Orchard and Yuchen Zhu for their valuable comments in a previous version of this article.
%We also thank Amazon for the financial support to Sergio under funding PSMEWE A03B “Merging data sources”.
\end{ack}

\newpage
\bibliography{main}

\newpage
\appendix
\section{Relation between the expectations of the Mixture of Gaussians and the known marginal expectations}
\label{app:relationGaussianConstraints}

In \Cref{th:anticausalDirection} we proved that the distribution resulting from the constraints in \Cref{eq:knownExpectations1,eq:knownExpectations2} and the anticausal optimisation problem in \Cref{eq:anticausalConditionalOptimisation} result in a mixture of Gaussian distributions.
Now we are going to explore the relation between the moments of the resulting distribution and the constraints used in the MAXENT optimisation problem.

We have the following expectations under Gaussian mixture model
\begin{align}
    \E\left[\bX Y\right] =& q\bmu_{1} - (1-q)\bmu_{-1}\\
    \E\left[\bX\right] =& q\bmu_{1} + (1-q)\bmu_{-1}\\
    \E\left[\bX\bX^{\top}\right] =& \bSigma_{X}\\
    =& \E\left[\Var(\bX\mid Y)\right] + \Var(\E\left[\bX\mid Y\right]).\label{eq:lawOfTotalCov}
\end{align}
Where \Cref{eq:lawOfTotalCov} follows from the law of total covariance. We have
\begin{align}
    \E\left[\Var(\bX\mid Y)\right] =& q\bSigma_{X\mid Y} + (1-q)\bSigma_{X\mid Y}\\
    =& \bSigma_{X\mid Y},\\
    \Var(\E\left[\bX\mid Y\right]) =& \E[\E[\bX\mid Y]^{2}] - \E[\E[\bX\mid Y]]^{2}\\
    \E[\E[\bX\mid Y]^{2}] =& q\bmu_1\bmu_1^\top+(1-q)\bmu_{-1}\bmu_{-1}^\top\\
    \E[\E[\bX\mid Y]]^{2} =& (q\bmu_1+(1-q)\bmu_{-1})(q\bmu_1 + (1-q)\bmu_{-1})^\top.
\end{align}
So that
\begin{align}
    \begin{split}
        \E\left[\bX\bX^{\top}\right] =& \bSigma_{X\mid Y} + q\bmu_1\bmu_1^\top+(1-q)\bmu_{-1}\bmu_{-1}^\top\\
        & - (q\bmu_1+(1-q)\bmu_{-1})(q\bmu_1 + (1-q)\bmu_{-1})^\top
    \end{split}\\
    =& \bSigma_{X\mid Y} + (1-q)q\bmu_1\bmu_1^\top + (1-q)q\bmu_{-1}\bmu_{-1}^\top - (1-q)q\bmu_{-1}\bmu_1^\top - (1-q)q\bmu_{1}\bmu_{-1}^\top\\
    =& \bSigma_{X\mid Y} + (1-q)q[\bmu_1\bmu_1^\top + \bmu_{-1}\bmu_{-1}^\top - \bmu_{-1}\bmu_1^\top - \bmu_{1}\bmu_{-1}^\top]\\
    =& \bSigma_{X\mid Y} + (1-q)q(\bmu_1-\bmu_{-1})(\bmu_1-\bmu_{-1})^{\top}\label{eq:conditionalCovariance}.
\end{align}
Recall that the empirical averages used as constraints in the maximum entropy optimisation problem are coincide with the expectations under the resulting exponential family distribution.
Then, using the equations above and the constraints, the means of the multivariate Gaussian distribution are
\begin{align}
    \bmu_{1} =& \frac{\barbx+\bphi}{2q}\label{eq:antiCausalMean1}\\
    \bmu_{-1} =& \frac{\barbx-\bphi}{2(1-q)}\label{eq:antiCausalMean-1}.
\end{align}
And the conditional covariance, which is the same for both components, is
\begin{align}
    \begin{split}
        \bSigma_{X\mid Y} =&
        \begin{bmatrix}
            \bars^{2}_{1} & \bars_{1,2}\\ 
            \bars_{1,2} & \bars^{2}_{2}
        \end{bmatrix}\\
        &- q(1-q) \left[\frac{(\barbx+\bphi)(\barbx+\bphi)^{\top}}{2^2(1-q)^2}
        + \frac{(\barbx-\bphi)(\barbx-\bphi)^{\top}}{2^2q^2}\right.\\
        &- \left.\frac{(\barbx+\bphi)(\barbx-\bphi)^{\top}}{2^2q(1-q)}
        - \frac{(\barbx-\bphi)(\barbx+\bphi)^{\top}}{2^2q(1-q)}\right]
    \end{split}\\
    \begin{split}
        \bSigma_{X\mid Y} =&
        \begin{bmatrix}
            \bars^{2}_{1} & \bars_{1,2}\\ 
            \bars_{1,2} & \bars^{2}_{2}
        \end{bmatrix}\\
        &- q(1-q) \left[\frac{\barbx\barbx^\top+\barbx\bphi^\top+\bphi\barbx^\top+\bphi\bphi^\top}{2^2(1-q)^2}
        +\frac{\barbx\barbx^\top-\barbx\bphi^\top-\bphi\barbx^\top+\bphi\bphi^\top}{2^2q^2}\right.\\
        &+\left.\frac{-\barbx\barbx^\top+\barbx\bphi^\top-\bphi\barbx^\top+\bphi\bphi^\top}{2^2q(1-q)}+
        \frac{-\barbx\barbx^\top-\barbx\bphi^\top+\bphi\barbx^\top+\bphi\bphi^\top}{2^2q(1-q)}\right]
    \end{split}\\    
    \begin{split}
        \bSigma_{X\mid Y} =&
        \begin{bmatrix}
            \bars^{2}_{1} & \bars_{1,2}\\ 
            \bars_{1,2} & \bars^{2}_{2}
        \end{bmatrix}\\
        &- \frac{q(1-q)}{2^2q^2(1-q)^2} 
        [\barbx\barbx^\top (q^2+(1-q)^2-q(1-q)-q(1-q))\\
        &+ \barbx\bphi^\top(q^2-(1-q)^2+q(1-q)-q(1-q))\\
        &+ \bphi\barbx^\top(q^2-(1-q)^2-q(1-q)+q(1-q))\\
        &+ \bphi\bphi^\top(q^2+(1-q)^2+q(1-q)+q(1-q))]
    \end{split}\\
    \begin{split}
        \bSigma_{X\mid Y} =&
        \begin{bmatrix}
            \bars^{2}_{1} & \bars_{1,2}\\ 
            \bars_{1,2} & \bars^{2}_{2}
        \end{bmatrix}\\
        &- \frac{1}{2^2q(1-q)} 
        [\barbx\barbx^\top (4q^2-4q+1)
        + \barbx\bphi^\top(2q-1)
        + \bphi\barbx^\top(2q-1)
        + \bphi\bphi^\top]
    \end{split}\\
    \begin{split}\label{eq:relationCovariances}
        \bSigma_{X\mid Y} =&
        \begin{bmatrix}
            \bars^{2}_{1} & \bars_{1,2}\\ 
            \bars_{1,2} & \bars^{2}_{2}
        \end{bmatrix}
        - \frac{1}{2^2q(1-q)} 
        [(2q-1)\barbx+\bphi][(2q-1)\barbx+\bphi]^\top
    \end{split}.
\end{align}
We enumerate the individual elements:
\begin{align}
\bSigma_{X\mid Y,\,(i,i)} =& \frac{1}{2^2q(1-q)} [\bars^{2}_{i}-(2q-1)^2\bar{x}_i^2-2(2q-1)\bar{x}_i\phi_i-\phi_i^2]\\
\bSigma_{X\mid Y,\,(1,2)} =& \frac{1}{2^2q(1-q)} [\bars_{1,2}-(2q-1)^2\bar{x}_1\bar{x}_2-(2q-1)\bar{x}_1\phi_2-(2q-1)\bar{x}_2\phi_1-\phi_1\phi_2].
\end{align}
\section{Predictor in the anticausal direction}

\bayesPredictor*
\begin{proof}
We prove this for the case $y=1$.
the case for $y=-1$ can be derived in an analogous way.
The result follows from the application of Bayes' rule:
\begin{align}
    \pdf&(y=1\mid\bx) = \frac{\pdf(\bx\mid y=1)\pdf(y=1)}{\pdf(\bx)}\\
    &= \frac{q\exp\left(-\frac{1}{2}(\bx-\bmu_{1})^{\top}\bSigma_{\bX\mid Y}^{-1}(\bx-\bmu_{1})\right)}
    {q\exp\left(-\frac{1}{2}(\bx-\bmu_{1})^{\top}\bSigma_{\bX\mid Y}^{-1}(\bx-\bmu_{1})\right) + (1-q)\exp\left(-\frac{1}{2}(\bx-\bmu_{-1})^{\top}\bSigma_{\bX\mid Y}^{-1}(\bx-\bmu_{-1})\right)}\\
    &= \frac{1}
    {1 + \frac{(1-q)}{q}\exp\left(-\frac{1}{2}(2\bx^{\top}\bSigma_{\bX\mid Y}^{-1}(\bmu_{1}-\bmu_{-1}) + \bmu^{\top}_{1}\bSigma_{\bX\mid Y}^{-1}\bmu_{1} - \bmu^{\top}_{-1}\bSigma_{\bX\mid Y}^{-1}\bmu_{-1}\right)}\\
    &= \frac{1}
    {1 + \frac{(1-q)}{q}\exp\left((\bx-\frac{\bmu_{-1}}{2})^{\top}\bSigma_{\bX\mid Y}^{-1}\bmu_{-1} - (\bx-\frac{\bmu_{1}}{2})^{\top}\bSigma_{\bX\mid Y}^{-1}\bmu_{1}\right)}
\end{align}
\end{proof}
\section{Derivation of the decision boundary}

In the following two sections we give the proof of \Cref{th:decisionBoundaryCMAXENT} for the causal and anticausal direction separately.
First, we restate the proposition
\decisionBoundaryCMAXENT*

As mentioned on the proposition, we frame these results within the statistical decision theory
framework \citep{berger2013statistical}, choosing a particular loss function $L(h(\bx), y)$,
where $h(\bx)$ is the predictor of $y$ we want to evaluate. 
We consider the 0-1 loss function. 
That is, $L(h(\bx), y)=1$ if $h(\bx)=y$, and 0 otherwise. 
The optimal decision rule for this loss is
the well-known Maximum A Posteriori (MAP) 
rule from which we can derive our decision boundary. 

\subsection{Proof of \Cref{th:decisionBoundaryCMAXENT} in the causal direction}
\label{ssec:decisionBoundaryCausal}

In the causal direction, the Maximum A Posteriori (MAP) rule, results in the decision boundary given by the following equation
\begin{align}
    \pdf(y=1\mid\bx) &= \pdf(y=-1\mid\bx)\\
    \frac{1}{2}(1+\tanh(\lambda_0 + \lambda_{1}x_1 + \lambda_{2}x_2)) &= \frac{1}{2}(1+\tanh(-\lambda_0 - \lambda_{1}x_1 - \lambda_{2}x_2))\\
    \lambda_0 + \lambda_{1}x_1 + \lambda_{2}x_2 &= -\lambda_0 - \lambda_{1}x_1 - \lambda_{2}x_2\\
    \lambda_0 + \lambda_{1}x_1 + \lambda_{2}x_2 &= 0.
\end{align}
In words, the decision boundary in the causal direction is a linear function of the covariates.
Using this result, we proceed to prove the relation between the marginal covariance matrix and the normal to the decision boundary as in \Cref{th:decisionBoundaryCausal} of \Cref{th:decisionBoundaryCMAXENT}

We want to prove $\blambda\propto \Sigma_{\bX}^{-1}\Sigma_{\bX,Y} = \Sigma_{\bX}^{-1}\bphi$.

First, we define the random variable $Z:= \lambda_1 X_1 + \lambda_2 X_2$.
We can write $\pdf(y=1 |\bx)$ entirely as function of $Z$, thus $\bX \indep Y\,|Z$.

To continue with the proof, we consider the Hilbert space of centered random variables with basis given by $\vecspan{(\bX)}$ and covariance as inner product. 
Following this geometric interpretation, we define $W_j:=X_j-\alpha_jZ$, where $\alpha_j Z$ is the projection of $X_j$ onto the span of $Z$.
That is, ${\alpha_j=\Cov[X_j, Z]\Var(Z)^{-1}}$.
We have that $\bW=\{W_1,W_2\}\in\vecspan{(\bX)}$, so that $\Cov[Z,\bW]=0$.
As a result, $\bW\indep Z$ because all variables in the span of $\bX$ are Gaussian. 
Together with $\bW\indep Y\mid Z$, this implies via the semi-graphoid axioms \citep{lauritzen1996graphical} that $\bW\indep (Y, Z)$, so that $\bW\indep Y$ and thus $\Cov[\bW,Y]=0$. 

Taking the inner product with $Y$ on both sides of $X_j=\alpha_jZ+W_j$ gives us $\Cov[X_j,Y] =\alpha_j\Cov[Z,Y]+\Cov[W_j,Y] =\alpha_j\Cov[Z,Y] =\Cov[X_j, Z]\Var(Z)^{-1}\Cov[Z,Y]$.
This is valid for $j=1,2$, hence $\Sigma_{\bX,Y}= \Sigma_{\bX, Z}\sigma_{Z}^{-2}\Sigma_{Z,Y}$.
By definition, $\Sigma_{\bX, Z}=\blambda\Sigma_{\bX}$, so that $\Sigma_{\bX,Y}= \blambda\Sigma_{\bX}\sigma_{Z}^{-2}\sigma_{Z,Y}$ and finally $(\Sigma_{\bX}\sigma_{Z}^{-2}\sigma_{Z,Y})^{-1}\Sigma_{\bX,Y} =\blambda$, as required.

\subsection{Proof of \Cref{th:decisionBoundaryCMAXENT} in the anticausal direction}
\label{ssec:decisionBoundaryAnticausal}

The normal to the decision boundary using the MAP rule in anticausal direction is derived in a similar way to \Cref{ssec:decisionBoundaryCausal}.
In particular, the normal is given by the points of $\bx$ where we are indifferent between choosing $y=1$ and $y=-1$.
To find such a vector, we solve for $\bx$ in
\begin{align}
    \pdf(y=1\mid\bx) &= \pdf(y=-1\mid\bx)\\
    \pdf(\bx\mid y=1)\pdf(y=1) &= \pdf(\bx\mid y=-1)\pdf(y=-1).
\end{align}
In the second line of the above equation, we used $\pdf(Y=y\mid \bx)\propto \pdf(\bx\mid Y=y)P(Y=y)$.

We have
\begin{align}
    q\exp\left(-\frac{1}{2}(\bx-\bmu_{1})^{\top}\bSigma_{\bX\mid Y}^{-1}(\bx-\bmu_{1})\right)
    = (1-q)\exp\left(-\frac{1}{2}(\bx-\bmu_{-1})^{\top}\bSigma_{\bX\mid Y}^{-1}(\bx-\bmu_{-1})\right)\\
    \log\left(\frac{q}{1-q}\right) =
    -\frac{1}{2}(2\bx^{\top}\bSigma_{\bX\mid Y}^{-1}(\bmu_{1}-\bmu_{-1}) + 
    \bmu^{\top}_{1}\bSigma_{\bX\mid Y}^{-1}\bmu_{1} - \bmu^{\top}_{-1}\bSigma_{\bX\mid Y}^{-1}\bmu_{-1}).
\end{align}

The above equation is linear in $\bx$, giving us a linear decision rule, and we would choose $Y=1$ if
\begin{align}\label{eq:anticausalEqualsLDA}
    \bx^{\top}\bSigma_{\bX\mid Y}^{-1}(\bmu_{1}-\bmu_{-1}) >&
    \frac{1}{2}\bmu^{\top}_{-1}\bSigma_{\bX\mid Y}^{-1}\bmu_{-1} -
    \frac{1}{2}\bmu^{\top}_{1}\bSigma_{\bX\mid Y}^{-1}\bmu_{1} -
    \log\left(\frac{q}{1-q}\right)\\
    =& \frac{1}{2}(\bmu_{-1}-\bmu_{1})^{\top}\bSigma_{\bX\mid Y}^{-1}(\bmu_{-1}+\bmu_{1}) -
    \log\left(\frac{q}{1-q}\right).
\end{align}

\begin{rmk}
    As mentioned in \Cref{th:bayesPredictor}, the decision rule in \Cref{eq:anticausalEqualsLDA} is known as the Gaussian discriminant analysis \citep{hastie2009elements}. 
    This is a special case of Linear Discriminant Analysis. 
    The family of LDA algorithms also contains Naive Bayes (if all the $\bx$ are conditionally independent) and Quadratic Discriminant Analysis (QDA) (if the covariance matrices for each $Y$ are not equal, giving a curved decision rule).
\end{rmk}

Now we will prove that if we have use all the moments in \Cref{eq:knownExpectations1,eq:knownExpectations2} as constraints, the slope of the two decision boundaries are the same.

\equalityOfSlopes*

\begin{proof}
    In \Cref{th:decisionBoundaryCMAXENT} we proved that in the causal direction, the normal vector to the decision boundary in the causal direction is $\Sigma_\bX^{-1}\bphi$. 
    Furthermore, using the law of total covariance (and the assumption that $\barx=0$), we can write $\bSigma_\bX=\bSigma_{\bX\mid Y}+c\bphi\bphi^\top$, where $c=1/(2^2q(1-q))$ (see \Cref{eq:relationCovariances}).
    Using the Sherman-Morrison formula \citep{bartlett1951inverse}, we can write 
    \begin{align}
        \bSigma_\bX^{-1} =& (\bSigma_{\bX\mid Y}+c\bphi\bphi^\top)^{-1}\\
        =& \bSigma_{\bX\mid Y}^{-1}-\frac{c\bSigma_{\bX\mid Y}^{-1}\bphi\bphi^\top\bSigma_{\bX\mid Y}^{-1}}
        {1+c\bphi^\top\bSigma_{\bX\mid Y}^{-1}\bphi}.
    \end{align}
    Applying this operator to $\bphi$, and noticing that $\bphi^\top\bSigma_{\bX\mid Y}^{-1}\bphi$ is a scalar, we obtain 
    \begin{align}
        \bSigma_\bX^{-1}=\bSigma_{\bX\mid Y}^{-1}\bphi+k\bSigma_{\bX\mid Y}^{-1}\bphi,
    \end{align}
    where $k=(-c\bphi^\top\bSigma_{\bX\mid Y}^{-1}\bphi)/(1+c\bphi^\top\bSigma_{\bX\mid Y}\bphi)$.
    Thus $\bSigma_\bX^{-1}\bphi\propto\bSigma_{\bX\mid Y}\bphi$, as required.
\end{proof}

\section{Missing covariance between the outcome variable and one of the covariates}

Suppose we do not observe $\E[YX_2]=\phi_2$.
Can CMAXENT say anything about $\pdf(Y\mid \bx)$?
The answer is positive under the assumption that $Y$ and $X_2$ are correlated.
To see this, we will use the following result in information theory:
The entropy of a distribution with given first and second moments is always less than the entropy of a multivariate Gaussian given the same first and second moments \citep[Theorem 8.6.5]{cover2006elements}.
Hence, we can analytically compute the maximum entropy solution of $\phi_2$ via the entropy of the multivariate Gaussian as an upper bound on the entropy of $\bX$ given $Y$.

To do this, we will use the results of \Cref{app:relationGaussianConstraints}, where we found an expression of $\bSigma_{\bX\mid Y}$ as a function of $\phi_2$ .
Since $\phi_2$ appears on several elements of the $\bSigma_{\bX\mid Y}$, we first compute the determinant of $\bSigma_{\bX\mid Y}$, differentiate with respect to $\phi_2$ and equate to 0 to find the optimal $\phi_2$.

For reference, the differential entropy of a multivariate Gaussian of $k$ dimensions and covariance matrix $\bSigma$ is
\begin{align}
    H(f) = \frac{k}{2}+\frac{k}{2}\log(2\pi)+\frac{1}{2}\log(\det(\bSigma))
\end{align}

First we compute the determinant of $\bSigma_{\bX\mid Y}$:
\begin{align}
\begin{split}
    \det(\bSigma_{X\mid Y}) =& \bars^2_1\bars^2_2 -\frac{\bars^2_1(2q-1)^2\barx^2_2-\bars^2_12(2q-1)\barx_2\phi_2-\bars^2_1\phi^2_2}{2q(1-q)}\\
    &- \frac{(2q-1)^2x^2_1\bars^2_2}{2q(1-q)}
    + \frac{(2q-1)^4\barx^2_1\barx^2_2+2(2q-1)^3\barx^2_1\barx_2\phi_2+(2q-1)^2\barx^2_1\phi^2_2}{(2q(1-q))^2}\\
    &- \frac{2(2q-1)\barx_1\phi_2\bars^2_2}{2q(1-q)}
    + \frac{2(2q-1)^3\barx_1\phi_1\barx^2_2+2^2(2q-1)^2\barx_1\barx_2\phi_1\phi_2+2(2q-1)\barx_1\phi_1\phi^2_2}{(2q(1-q))^2}\\
    &- \frac{\phi^2_1\bars^2_2}{2q(1-q)}
    + \frac{(2q-1)\barx^2_2\phi^2_1+2(2q-1)\barx_2\phi^2_1\phi_2+\phi^2_1\phi^2_2}{(2q(1-q))^2}\\
    &- \bars^2_{1,2} 
    + \frac{\bars_{1,2}(2q-1)^2\barx_1\barx_2+\bars_{1,2}(2q-1)\barx_1\phi_2+\bars_{1,2}(2q-1)\barx_2\phi_1+\bars_{1,2}\phi_1\phi_2}{2q(1-q)}\\
    &+ \frac{(2q-1)^2\barx_1\barx_2\bars_{1,2}}{2q(1-q)}
    - \frac{(2q-1)^4\barx^2_1\barx^2_2-(2q-1)^3\barx^2_1\barx_2\phi_2-(2q-1)^3\barx_1\barx^2_2\phi_1}{(2q(1-q))^2}\\
    &- \frac{(2q-1)^2\barx_1\barx_2\phi_1\phi_2}{(2q(1-q))^2}
    + \frac{(2q-1)\barx_1\phi_2\bars_{1,2}}{2q(1-q)}
    - \frac{(2q-1)^3\barx^2_1\phi_2\barx_2-(2q-1)^2\barx^2_1\phi^2_2}{(2q(1-q))^2}\\
    &- \frac{(2q-1)^2\barx_1\barx_2\phi_1\phi_2-(2q-1)\barx_1\phi_1\phi^2_2}{(2q(1-q))^2}
    + \frac{(2q-1)\barx_2\phi_1\bars_{1,2}}{2q(1-q)}\\
    &- \frac{(2q-1)^3\barx_1\barx^2_2\phi_1-(2q-1)^2\barx_1\barx_2\phi_1\phi_2}{(2q(1-q))^2}
    - \frac{(2q-1)^2\barx^2_2\phi^2_1-(2q-1)\barx_2\phi^2_1\phi2}{(2q(1-q))^2}\\
    &+ \frac{\phi_1\phi_2\bars_{1,2}}{2q(1-q)} 
    - \frac{(2q-1)^2\barx_1\barx_2\phi_1\phi_2-(2q-1)\barx_1\phi_1\phi^2_2}{(2q(1-q))^2}
    - \frac{(2q-1)\barx^2_2\phi^2_1\phi_2-\phi^2_1\phi^2_2}{(2q(1-q))^2}.
\end{split}
\end{align}

Now we differentiate $\det\left(\bSigma_{X\mid Y}\right)$ with respect to $\phi_2$, equate to 0 and solve for $\phi_2$
\begin{align}
\begin{split}
    \frac{\partial\det\bSigma_{X\mid Y}}{\partial\phi_2} =& 
    - \frac{\bars^2_12(2q-1)\barx_2}{2q(1-q)}
    - \frac{2\bars^2_1\phi_2}{2q(1-q)}
    + \frac{2(2q-1)^3\barx^2_1\barx_2}{(2q(1-q))^2}
    + \frac{2(2q-1)^2\barx^2_1\phi_2}{(2q(1-q))^2}\\
    &+\frac{2^2(2q-1)^2\barx_1\barx_2\phi_1}{(2q(1-q))^2}
    + \frac{2^2(2q-1)\barx_1\phi_1\phi_2}{(2q(1-q))^2}
    + \frac{2(2q-1)\barx_2\phi_1^2}{(2(2q-1))^2}
    + \frac{2\phi_1^2\phi_2}{(2(2q-1))^2}\\
    &+ \frac{\bars_{1,2}(2q-1)\barx_1}{2q(1-q)}
    + \frac{\bars_{1,2}\phi_1}{2q(1-q)}
    - \frac{(2q-1)^3\barx_1^2\barx_2}{(2(2q-1))^2}
    - \frac{(2q-1)^2\barx_1\barx_2\phi_1}{(2(2q-1))^2}\\
    &+ \frac{(2q-1)\barx_1\bars_{1,2}}{2q(1-q)}
    - \frac{(2q-1)^3\barx_1^2\barx_2}{(2q(1-q))^2}
    - \frac{2(2q-1)^2\barx_1^2\phi_2}{(2q(1-q))^2}
    - \frac{(2q-1)^2\barx_1\barx_2\phi_1}{(2q(1-q))^2}\\
    &- \frac{2(2q-1)\barx_1\phi_1\phi_2}{(2q(1-q))^2}
    - \frac{(2q-1)^2\barx_1\barx_2\phi_1}{(2q(1-q))^2}
    - \frac{(2q-1)\barx_2\phi_1^2}{(2q(1-q))^2}.
\end{split}
\end{align}

Equating the above derivative to 0, we obtain
\begin{align}
\begin{split}
    \phi_2 &[-2\bars_1^2q(1-q) +2(2q-1)^2\barx_1^2 +2^2(2q-1)\barx_1\phi_1 +2\phi_1^2 -2(2q-1)^2\barx_1^2 -2(2q-1)\barx_1\phi_1]\\
    &= \bars_1^22^22q(1-q)(2q-1)\barx_2 -2(2q-1)^3\barx_1^2\barx_2 -2^2(2q-1)^2\barx_1\barx_2\phi_1\\
    &- 2(2q-1)\barx_2\phi_1^2 -\bars_{1,2}2q(1-q)(2q-1)\barx_1 -2q(1-q)\bars_{1,2}\phi_1\\
    &+ (2q-1)^3\barx_1^2\barx_2 +(2q-1)^2\barx_1\barx_2\phi_1 -(2q-1)\barx_1\bars_{1,2}\\
    &+ (2q-1)^3\barx_1^2\barx_2 +(2q-1)^2\barx_1\barx_2\phi_1 +(2q-1)2\barx_1\barx_2\phi_1 +(2q-1)\barx_2\phi_1^2.
\end{split}
\end{align}

Which can be simplified to
\begin{align}
\begin{split}
    \phi_2 =& \frac{1}{2\phi_1^2+2(2q-1)\barx_1\phi_1-2\bars_1^2q(1-q)}\cdot\\
    &[2^2q(1-q)(2q-1)\bars_1^2\barx_2- (2q-1)\barx_2\phi_1^2- 2q(1-q)(2q-1)\bars_{1,2}\barx_1\\
    &- 2q(1-q)\bars_{1,2}\phi_1- (2q-1)\barx_1\bars_{1,2}+ (2q-1)^2q(1-q)]. 
\end{split}
\end{align}

Although we have derived here the general case where the sample means are not zero, we will continue the analysis by coming back to such assumption.
That is, $\barx_1=\barx_2=0$, giving us the following expression for $\phi_2$ which is easier to interpret
\begin{align}
    \phi_2 =& \frac{q(1-q)\bars_{1,2}\phi_1}{q(1-q)\bars_1^2-\phi_1^2}.
\end{align}

In the numerator, we have that as the covariance between $X_1$ and $X_2$ increases, the MAXENT covariance between $X_2$ and $Y$ increases too. 
Furthermore, we see that the denominator is always greater than 1 by the Cauchy-Schwarz inequality of random variables, $\Cov(X_1,Y)^2\leq \Var(X_1)\Var(Y)$, given that $q(1-q)$ is the variance of $Y$, $\bars_1^2$ is the sample variance of $X_1$, and $\phi_1$ is the sample covariance between $X_1$ and $Y$.
\section{Derivation of the decision boundary with unknown predictor covariance} \label{app:unknownPredictorCovariance}

\paragraph{Causal.}In the causal case, we have that the distribution of the causes is a multivariate Gaussian with diagonal covariance matrix, and the conditional distribution of the target variable given the covariates is the same as in \Cref{eq:causalConditionalMaxent}.

As a result, we have that the decision boundary is still proportional to 
\begin{align}\label{eq:decisionBoundaryNoCovarianceCausal}
    \bSigma_{\bX,causal}^{-1}\bphi= % Add another equality with the explicity expression for \bSigma_{\bX,causal}^{-1} being a diagonal matrix.
    \begin{bmatrix}
        \bars^{-2}_1\phi_1\\
        \bars^{-2}_2\phi_2
    \end{bmatrix}
\end{align}
as derived in section \Cref{ssec:decisionBoundaryCausal}.

\paragraph{Anticausal.}In the anticausal direction, we have that the target variable follows a Bernoulli distribution, and the conditional distribution of the covariates given the target variable is again a mixture of Gaussians with diagonal conditional covariance matrix. 
We first prove \Cref{th:unknownCovarianceAnticausalDiagonalConditionalCovariance}.
\unknownCovarianceAnticausalDiagonalConditionalCovariance*
\begin{proof}
    The solution to the constrained optimisation problem has the same form as in \Cref{eq:exponentialAnticausal}, without the cross term:
    \begin{align}
        \pdf_\lambda(\bx \mid y) 
        = \exp[\lambda_{1}yx_{1} + \lambda_{2}yx_{2} 
        + \lambda_{3}x_{1} + \lambda_{4}x_{2}
        + \lambda_{5}x^{2}_{1} + \lambda_{6}x^{2}_{2} + \beta(y)].
    \end{align}
    Conditioning on any specific value of $Y$, gives us an uncorrelated multivariate Gaussian, as required.
\end{proof}

Using \Cref{eq:conditionalCovariance} we can express the conditional covariance as
\begin{align}
    \bSigma_{\bX\mid Y} =& \bSigma_{\bX} - (1-q)q(\bmu_1-\bmu_{-1})(\bmu_1-\bmu_{-1})^{\top}\\
    =& \begin{bmatrix}
        \bars_1^2 & \psi\\
        \psi & \bars_2^2
    \end{bmatrix}
    - q(1-q)
    \begin{bmatrix}
        (\mu_{1,1}-\mu_{-1,1})^2 & (\mu_{1,1}-\mu_{-1,1})(\mu_{1,2}-\mu_{-1,2})\\
        (\mu_{1,2}-\mu_{-1,2})(\mu_{1,1}-\mu_{-1,1}) & (\mu_{1,2}-\mu_{-1,2})^2
    \end{bmatrix}.
\end{align}
Since we know that $\bSigma_{\bX\mid Y}$ is diagonal, then $\psi=q(1-q)(\mu_{1,1}-\mu_{-1,1})(\mu_{1,2}-\mu_{-1,2})$. 
From \Cref{eq:antiCausalMean1,eq:antiCausalMean-1}, we can conclude that $q(1-q)(\mu_{1,i}-\mu_{-1,i})^2\propto \phi_i^2$.
Wit this, we find an expression of $\bSigma_{\bX\mid Y}$ as a function of the constraints
\begin{align}
    \bSigma_{\bX\mid Y} =
    \begin{bmatrix}
        \bars_1^2-\phi_1^2 & 0\\
        0 & \bars_2^2-\phi_2^2
    \end{bmatrix}.
\end{align}
On \Cref{ssec:decisionBoundaryAnticausal} (see also \citet[Sec. 4.4.5]{hastie2009elements}) we proved that the slope of the decision boundary in the anticausal direction is proportional to 
\begin{align}
    \bSigma_{\bX\mid Y}^{-1}\bphi,
\end{align}
and using \cref{eq:antiCausalMean1,eq:antiCausalMean-1}, we have that the slope of the decision boundary is proportional to
\begin{align}\label{eq:decisionBoundaryNoCovarianceAnticausal}
    \bSigma_{\bX\mid Y}^{-1}\bphi \propto
    \begin{bmatrix}
        (\bars_1^2-\phi_1^2)^{-1}\phi_1\\
        (\bars_2^2-\phi_2^2)^{-1}\phi_2
    \end{bmatrix}.
\end{align}
A natural question arises: when are these slopes the same? 
in other words, when are \cref{eq:decisionBoundaryNoCovarianceCausal,eq:decisionBoundaryNoCovarianceAnticausal} linearly dependent?
This question can be answered be equating 
\begin{align}
    \frac{(\bars_1^2-\phi_1^2)^{-1}\phi_1\bars_2^2\phi_2}{\bars_1^2\phi_1},
\end{align}
to
\begin{align}
    (\bars_2^2-\phi_2^2)^{-1}\phi_2.
\end{align}
We have
\begin{align}
    &\frac{(\bars_1^2-\phi_1^2)^{-1}\phi_1\bars_2^2\phi_2}{\bars_1^2\phi_1} = (\bars_2^2-\phi_2^2)^{-1}\phi_2 &&\iff\\
    &(\bars_2^2-\phi_2^2)\phi_1\bars_2^2\phi_2 = (\bars_1^2-\phi_1^2)\phi_2\bars_1^2\phi_1 &&\iff\\
    &\frac{(\bars_2^2-\phi_2^2)\phi_1\bars_2^2\phi_2}{(\bars_1^2-\phi_1^2)\phi_2\bars_1^2\phi_1} = 1&&\iff\\
    &\frac{(\bars_2^2-\phi_2^2)\bars_2^2}{(\bars_1^2-\phi_1^2)\bars_1^2} = 1.
\end{align}

\section{Derivation of the target predictor when merging predictors in causal and anticausal direction} \label{app:causal-and-anticausal-predictor}

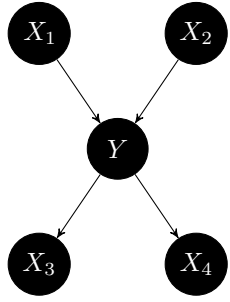
\begin{figure}
     \centering
     %!TEX root = CausalityCausalAndAnticausalMergingOfPredictors.tex

\begin{tikzpicture}
\node(Ref) {};
\node[obs,left=.5cm of Ref] (X1) {$X_1$};
\node[obs,right=.5cm of Ref] (X2) {$X_2$};
\node[obs,below=1cm of Ref] (Y) {$Y$};
\node[below=1cm of Y] (Ref2) {};
\node[obs,left=.5cm of Ref2] (X3) {$X_3$};
\node[obs,right=.5cm of Ref2] (X4) {$X_4$};

\edge{X1}{Y};
\edge{X2}{Y};
\edge{Y}{X3};
\edge{Y}{X4};

\end{tikzpicture}
     \caption{Graph in the causal and anticausal direction}
     \label{fig:graphCausalAndAnticausalDirection}
\end{figure}

In this section we explore the predictor resulting from merging predictors including causes and predictors including effects of the target variable.
We will assume the causal graph in \Cref{fig:graphCausalAndAnticausalDirection}.
We will use the first and second moments of each variable, the covariance between each $X_i$ and $Y$, the covariance between $X_1$ and $X_2$, and between $X_3$ and $X_4$, as constraints.

Using CMAXENT, we first find the density $\pdf(X_1,X_2)$ with maximum entropy subject to the moment constraints;
then $\pdf(Y\mid X_1,X_2)$ with maximum entropy subject to the moment constraints (including $\pdf(X_1,X_2)$, found in the previous step);
and finally the density $\pdf(X_3,X_4\mid Y)$ that maximises the entropy subject to the moment constraints (again, including the found $\pdf(Y)$).

It is possible to see that these process will result in the same predictors as in \Cref{sec:knownPredictorCovariances}.
That is, we find that $\pdf(X_1,X_2)$ is a multivariate Gaussian, $\pdf(Y\mid X_1,X_2)$ a logistic-like regression, and $\pdf(X_3,X_4\mid Y)$ a Mixture of Bivariate Gaussians.

These distributions provide us with enough information to find the joint distribution of all our variables $\pdf(Y,X_1,X_2,X_3,X_4)$, with which we can derive our predictor of interest.
We have
\begin{align}
\pdf(y\mid x_1,x_2,x_3,x_4)
=& \frac{\pdf(x_1,x_2,x_3,x_4\mid y)\pdf(y)}{\pdf(x_1,x_2,x_3,x_4)}\\
=& \frac{\pdf(x_1,x_2\mid y)\pdf(x_3,x_4\mid y)\pdf(y)}{\pdf(x_1,x_2,x_3,x_4)}\\
=& \frac{\pdf(y\mid x_1,x_2)\pdf(x_1,x_2)\pdf(x_3,x_4\mid y)\pdf(y)}{\pdf(y)\pdf(x_1,x_2,x_3,x_4)}\\
=& \frac{\pdf(y\mid x_1,x_2)\pdf(x_1,x_2)\pdf(x_3,x_4\mid y)}{\pdf(x_1,x_2,x_3,x_4)}\\
=& \frac{\pdf(y\mid x_1,x_2)\pdf(x_1,x_2)\pdf(x_3,x_4\mid y)}{\sum_y \pdf(x_1,x_2,x_3,x_4\mid y)\pdf(y)}\\
=& \frac{\pdf(y\mid x_1,x_2)\pdf(x_1,x_2)\pdf(x_3,x_4\mid y)}{\sum_y \pdf(y\mid x_1,x_2)\pdf(x_1,x_2)\pdf(x_3,x_4\mid y)}.\label{eq:predictor-causal-and-anticausal-direction}
\end{align}
Where the second inequality follows from the conditional independence between $\set{X_1,X_2}$ and $\set{X_3,X_4}$ given $Y$, and the third inequality follows from Bayes' rule.

\Cref{eq:predictor-causal-and-anticausal-direction} gives us the desired predictor.
Notice that we found all of the elements needed to compute $p(y\mid \bx)$ on \Cref{sec:knownPredictorCovariances}.

%%%%%%%%%%%%%%%%%%%%%%%%%%%%%%%%%%%%%%%%%%%%%%%%%%%%%%%%%%%%
%\input{sections/checklist}

\end{document}